%% file: 00-main.tex
\pdfoutput=1

\documentclass[11pt]{article}

\usepackage[]{acl}

\usepackage{times}
\usepackage{latexsym}

\usepackage[T1]{fontenc}

\usepackage[utf8]{inputenc}

\usepackage[russian,arabic,farsi,english]{babel}

\usepackage{microtype}
\usepackage{booktabs}
\usepackage{multirow}
\usepackage{scalerel}
\usepackage{tabularx}
\usepackage{subcaption}

\title{What about Zir? On Pronouns in Machine Translation}
\title{What about Zir? \\ How Commercial Machine Translation Systems Fail to Handle Pronouns}
\title{What about \emph{Zir}? \\ How Commercial Machine Translation Fails to Handle Pronouns}
\title{What about \emph{em}? \\ How Commercial Machine Translation Fails to Handle (Neo-)Pronouns}

\newcommand{\al}[1]{\textcolor{black}{#1}}

\newcommand{\final}[1]{\textcolor{black}{#1}}

\usepackage{tabularx}

\newcolumntype{a}{>{\hsize=.6\hsize}X}

  \author{Anne Lauscher\textsuperscript{1}, Debora Nozza\textsuperscript{2}, Archie Crowley\textsuperscript{3}, Ehm Miltersen\textsuperscript{4}, and Dirk Hovy\textsuperscript{2} \\
    \textsuperscript{1}Data Science Group, Universität Hamburg, Germany \\
  \textsuperscript{2}Department of Computing Sciences, Bocconi University, Italy \\
  \textsuperscript{3}Linguistics, University of South Carolina \\
  \textsuperscript{4}School of Culture and Communication, Aarhus University \\
  \small{
  \texttt{anne.lauscher@uni-hamburg.de}, \texttt{\{debora.nozza, dirk.hovy\}@unibocconi.it}}, \\ \small{\texttt{acrowley@sc.edu},  \texttt{e.hjorth.miltersen@gmail.com}}}

\begin{document}
\maketitle
\begin{abstract}
As 3rd-person pronoun usage shifts to include novel forms, e.g., neopronouns, we need more research on identity-inclusive NLP. Exclusion is particularly harmful in one of the most popular NLP applications, machine translation (MT). Wrong pronoun translations can discriminate against marginalized groups, e.g., non-binary individuals~\final{~\citep{dev-etal-2021-harms}}. In this ``reality check'', we study how three commercial MT systems translate 3rd-person pronouns. Concretely, we compare the translations of gendered vs.\ gender-neutral pronouns from English to five other languages (Danish, Farsi, French, German, Italian), and vice versa, from Danish to English.
Our error analysis shows that the presence of a gender-neutral pronoun often leads to grammatical and semantic translation errors. Similarly, gender neutrality is often not preserved. By surveying the opinions of affected native speakers from diverse languages, we provide recommendations to address the issue in future MT research. 
\end{abstract}

\section{Introduction}
\input{01-intro}

\section{Related Work}
\input{02-rw}

\section{The Status Quo}
\input{04-from-en}

\section{What Would Be a Good Translation?}
\input{05-survey}

\section{Conclusion}
\input{07-conclusion.tex}

\section*{Acknowledgements}
Part of this work is funded by the European Research Council (ERC) under the European Union’s Horizon 2020 research and innovation program (grant agreement No. 949944, INTEGRATOR). Anne Lauscher's work is funded under the Excellence Strategy of the Federal Government and the Länder. Debora Nozza and Dirk Hovy are members of the MilaNLP group and the Data and Marketing Insights Unit of the Bocconi Institute for Data Science and Analysis.

\section*{Limitations}
\input{0x-limitations.tex}

\section*{Ethics Statement}
\input{0xx-ethics.tex}

\bibliography{custom}
\bibliographystyle{acl_natbib}

\appendix

\input{0xxx-appendix.tex}

\end{document}

%% file: 01-intro.tex
Machine translation (MT) is one of the most common applications of NLP, with millions of daily users interacting with popular commercial providers (e.g., Bing, DeepL, or Google Translate). 
Given MT's widespread use and the increased focus on fairness in language technologies~\citep[e.g.,][]{hovy-spruit-2016-social, blodgett-etal-2020-language}, previous work has pointed to the potential ethical issues stemming from stereotypical biases encoded in the models, e.g., gender or age bias~\citep[e.g.,][\emph{inter alia}]{stanovsky-etal-2019-evaluating,levy-etal-2021-collecting-large}. 

Still, these studies treat gender as a binary variable and ignore the larger spectrum of (possibly marginalized) identities, e.g., non-binary individuals. This gender exclusivity stands in stark contrast to the findings of \citet{dev-etal-2021-harms}. Their survey of queer individuals showed that MT has the most potential for representational and allocational harms~\citep{barocas2017problem} for non-cis users (compared to other NLP applications). In this context, survey respondents mentioned the translation of \emph{pronouns} as particularly sensitive, as gender-neutral pronouns might be translated into gendered pronouns, resulting in harmful misgendering.

While individual studies have investigated the translation of established (gender-neutral) pronouns \cite[e.g., from Korean to English;][]{cho-etal-2019-measuring}, NLP research, in general, has ignored the \emph{``modern world of pronouns''} as recently described by \newcite{lauscher2022welcome}. They discuss the large variety of existing phenomena in English 3rd-person pronoun usage, with more traditional neopronoun sets (e.g., \emph{xe/xem})\footnote{\final{Throughout this work, we use the expression ``traditional neopronoun'' to refer to sets that are, in contrast to only recently described phenomena (e.g., nounself pronouns), already academically discussed for longer.}} and novel pronoun-related phenomena \cite[e.g., nounself pronouns like  \emph{vamp/vamp};][]{miltersen2016nounself}, which possibly match distinct aspects of an individuals identity. %

As an example of ubiquitous NLP technology, \textbf{truly inclusive MT should account for linguistic varieties that express identity aspects}, like the large spectrum of pronouns related to the social push to respect diverse identities. However, until now, (a) there has been no information on how our systems (fail to) handle this linguistic shift, and (b) it is unclear how MT should deal with novel pronouns. This case is especially challenging when source language pronouns do not have direct correspondences in the target language.

\vspace{0.3em}
\noindent\textbf{Contributions.} 
In this ``reality check'', we investigate the handling of various (neo)pronouns in MT for advancing inclusive NLP. To this end, we combine an extensive analysis of MT performance across six languages (Danish, English, Farsi, French, German, and Italian) and three commercial MT engines (Bing, DeepL, and Google Translate) with results from the largest survey on pronoun usage among queer individuals in AI to date. We answer the following four research questions (\textbf{RQs}): 

\vspace{0.4em}
\noindent\textbf{(RQ1)} \emph{How do gender-neutral pronouns affect the overall translation quality?} We show that compared to gendered pronouns, the translated output's grammaticality and the source sequence's semantic consistency \textbf{drops by up to 16 percentage points and 47 percentage points}, respectively, for some categories of neopronouns.

\vspace{0.4em}
\noindent\textbf{(RQ2)} \emph{How do MT engines handle gender-neutral pronouns?} We demonstrate that the strategies for how MT engines handle pronouns vary by pronoun category: while gendered pronouns are most often translated (89\%), engines tend to simply copy some categories of neopronouns (e.g., 74\% for the category of numberself-pronouns). %

\vspace{0.4em}
\noindent\textbf{(RQ3)} \emph{Which MT strategies for handling gender-neutral pronouns ``work''?} We show that in 56\% of cases when a traditional neopronoun is translated, it is translated to a gendered pronoun in the target language, \textbf{likely leading to misgendering}.

\vspace{0.4em}
\noindent\textbf{(RQ4)} \emph{How should MT handle pronouns?} The answers of 49 participants (149 participants in the pre-study) in our survey reflect the diversity of pronoun choices across English and other languages and the diversity of preferences in how individuals' pronouns should be handled. There is no clear consensus! We thus recommend providing \textbf{configuration options to adjust the treatment of pronouns to individuals' needs}. %

%% file: 02-rw.tex
We review works on gender bias in MT and the broader area of (gender)
identity inclusion in NLP. For a thorough survey on gender bias in MT, we refer to \citep{savoldi2021gender}.

\paragraph{Gender Bias in MT.}  
As with other areas of NLP~\citep[e.g.,][\emph{inter alia}]{bolukbasi2016man,gonen2019lipstick,DEBIE, barikeri-etal-2021-redditbias}, much research has been conducted on assessing (binary) gender bias in MT.
Most prominently, \citet{stanovsky-etal-2019-evaluating} presented the WinoMT corpus, which allows for assessing occupational gender bias as an extension of Winogender~\citep{rudinger-etal-2018-gender} and WinoBias ~\citep{zhao-etal-2018-gender}. \citet{troles2021extending} further extended WinoMT with gender-biased verbs and adjectives. Those corpora are template-based, while \citet{levy-etal-2021-collecting-large} focused on collecting natural data, and \citet{gonen2020automatically} proposed an automatic approach to detect gender issues in real-world input. \final{\citet{renduchintala-etal-2021-gender} analyzed the effect of efficiency optimization on the measurable gender bias.}  Focusing on a different perspective, \citet{hovy-etal-2020-sound} assessed stylistic (gender) bias in translations. Other studies have examined specific language pairs, e.g., English and Hindi~\citep{ramesh-etal-2021-evaluating}, English and Italian~\citep{vanmassenhove2021gender}, or English and Turkish~\citep{ciora-etal-2021-examining}. Similarly, \citet{cho-etal-2019-measuring} studied English--Korean translations focusing  on translating gender-neutral pronouns from Korean. They introduced a  measure reflecting the preservation of gender neutrality but do not consider any neopronouns. 
Based on similar data sets and measures, researchers have also addressed gender bias in MT, e.g., via domain adaptation~\citep{saunders-byrne-2020-reducing}, debiasing representations~\citep{escude-font-costa-jussa-2019-equalizing}, adding contextual information~\citep{basta-etal-2020-towards}, and training on gender-balanced corpora~\citep{costa-jussa-de-jorge-2020-fine}. Some mitigation approaches exploit explicit gender annotations to guide the model in choosing the intended gender~\citep[e.g.,][]{stafanovics-etal-2020-mitigating}. In this context, \citet{saunders-etal-2020-neural} proposed a schema for adding inflection tags. For instance, they demonstrated how gender-neutral entities can be translated from English to another language by using a non-binary inflection tag.

\paragraph{Gender and Identity-Inclusion in NLP.} 
While most MT studies on gender bias deal with a binary notion of gender, researchers have started to study non-binary gender and identity inclusivity in NLP downstream tasks and models.
\citet{qian2022perturbation} explored the robustness of models to demographic change using a perturber model that also considers non-binary gender identities, \citet{cao-daume-iii-2020-toward} studied gender inclusion in co-reference resolution, and \citet{brandl-etal-2022-conservative} analyzed how gender-neutral pronouns are handled by language models in Danish, English, and Swedish for natural language inference and co-reference resolution. \final{\citet{nozza-etal-2022-measuring} and \citet{holtermann-etal-2022-fair} measured bias and harmfulness in language models towards LGBTQIA+ individuals.} Other researchers focused on the problem more broadly. \citet{orgad-belinkov-2022-choose} mention the binary treatment of gender as one of the essential pitfalls in gender bias evaluation, and \citet{dev-etal-2021-harms} surveyed the harms arising from non-binary exclusion in NLP, indicating MT as one particularly harmful application. Following up, \citet{lauscher2022welcome} explored the various phenomena related to 3rd-person pronoun usage in English, e.g., neopronouns. %
We are the first to study the translation of these novel pronoun-related phenomena in MT.

%% file: 04-from-en.tex
\label{sec:experiment}
To shed light on the state of identity inclusion through 3rd person pronouns in commercial MT, we conduct a thorough error analysis when translating from English (\textsc{en}) to five diverse languages. We further describe an experiment opposite to this, translating from Danish (\textsc{da}) to \textsc{en}, in \S\ref{sec:to_en}.

\subsection{Experimental Setup}
Our overall setup consists of 3 steps: (1)~we create \textsc{en} source sentences, each of which contains %
3rd person pronouns representing different ``pronoun categories'' (e.g., \emph{gendered pronoun}, etc.) in different grammatical cases. (2) Next, we employ an MT system to translate the \textsc{en} sentences to five target languages. (3) Last, we let native speakers manually analyze the translations with respect to diverse criteria, e.g., \emph{grammaticality of the output}.

\paragraph{Creation of \textsc{en} Source Data.} 
\setlength{\tabcolsep}{4pt}
\begin{table*}[t]\centering
\small
\begin{tabularx}{\linewidth}{ll}\toprule
\textbf{Case} & \textbf{Templates}   \\
\midrule
\multirow{4}{*}{Nominative} & \emph{The attendant recognized the developer because <n> reads a lot of technical news.} \\
& \emph{The analyst employed the housekeeper because <n> could not stand housework.} \\
& \emph{Olivia lost the game, so <n> was sad.} \\
& \emph{Liam received a good grade, so <n> was happy.} \\
\cmidrule(lr){2-2}
\multirow{4}{*}{Accusative}	& \emph{The developer wanted free bread from the baker and made up a story for <a> about not having a kitchen.}	\\
& \emph{The attendant did not want to fight with the guard and gave <a> flowers.} \\
& \emph{I like Olivia, so I met <a> today.} \\
& \emph{I do not like Liam, so I do not want to meet <a> today.} \\
\cmidrule(lr){2-2}
\multirow{4}{*}{Poss. Depen.}	& \emph{The mechanic visited the writer and helped on fixing <pd> car engine.}	\\
& \emph{The baker sold bread to the CEO and enjoyed <pd> visits.} \\
& \emph{Liam lost <pd> phone.}		\\
&	\emph{Olivia found <pd> ring.}	\\
\cmidrule(lr){2-2}
\multirow{4}{*}{Poss. Indep.} & \emph{During lunch, the janitor looked for the attendant to steal <pi>.} \\ 
& \emph{Last Saturday, the physician called the tailor to fix <pi>.} \\
& \emph{I had no phone, so Olivia gave me <pi>.} \\
& \emph{I lost my notes, so Liam gave me <pi>.} \\
\cmidrule(lr){2-2}
\multirow{4}{*}{Reflexive} & \emph{The farmer did not want to talk to the writer because <n> was burying <r> in writing a new novel.} \\
& \emph{The chief employed the receptionist because <n> was too busy to answer those phone calls by <r> every day.} \\
& \emph{Olivia wanted to impress, so <n> baked a cake <r>.} \\
&\emph{Liam wanted a new haircut, so <n> cut the hair <r>.} \\
\bottomrule
\end{tabularx}
\caption{The templates we use for each grammatical case. Placeholders are indicated with brackets and the grammatical case of the pronoun to fill, e.g., \emph{<pd>} (possessive dependent pronoun). \final{The first two templates for each case are extracted from WinoMT~\citep{stanovsky-etal-2019-evaluating}, while the second two templates are added by us.} }\label{tab:templates}
\end{table*}
We start with the WinoMT data set~\citep{stanovsky-etal-2019-evaluating}, designed to assess gender bias in MT and consisting of sentences that contain occupations stereotypically associated with women (e.g., \emph{secretary}) or men (e.g., \emph{developer}). We conduct an automatic morphological analysis on each pronoun in the data set.\footnote{\final{For this, we use part-of-speech tags and morphological output generated by spaCy (\url{https://spacy.io})}.} Based on the output, we randomly sample for each grammatical case (e.g., nominative, etc.),  in which a 3rd person pronoun referring to an occupation appears in,  
two sentences: one in which the target occupation is stereotypically associated with men and one in which it is stereotypically associated with women. We then replace those pronouns with placeholders, indicating the case (e.g., \emph{<n>} for nominative) of each. Since WinoMT does not contain pronouns in the \emph{possessive independent} case, we create these by sampling additional sentences with \emph{possessive dependent} pronouns and remove the target noun. Accordingly, we end up with 10 templates from WinoMT (2 for each of the 5 grammatical cases). Additionally, given that WinoMT sentences are designed to be more complex and ambiguous, we manually create two additional, simpler sentences for each grammatical case (10 in total). In these sentences, the pronoun placeholders refer to given names. In accordance with the WinoMT pattern, we choose the top name stereotypically associated with women and the top name stereotypically associated with men according to 2020 U.S. Social Security name statistics.\footnote{\url{https://www.ssa.gov/oact/babynames/}} 
We show example templates in Table \ref{tab:templates}.

\setlength{\tabcolsep}{4pt}
\begin{table}[t]\centering
\small
\begin{tabularx}{\linewidth}{llllll}\toprule
\textbf{Phenomon} & \textbf{N} & \textbf{A} & \textbf{PD} & \textbf{PI} & \textbf{R} \\
\midrule
\multirow{2}{*}{Gendered} & \emph{he} & \emph{him} & \emph{his} & \emph{his} & \emph{himself} \\
& \emph{she} & \emph{her} & \emph{her} & \emph{hers} & \emph{herself} \\
\multirow{2}{*}{Gender-neutral} & \multirow{2}{*}{\emph{they}} & \multirow{2}{*}{\emph{them}} & \multirow{2}{*}{\emph{their}} & \multirow{2}{*}{\emph{theirs}} & \emph{themselves}  \\
&&&&& \emph{themself} \\
\multirow{2}{*}{Neo} & \emph{xe} & \emph{xem} & \emph{xyr} & \emph{xyrs} & \emph{xemself} \\
 & \emph{ey} & \emph{em} & \emph{eir} & \emph{eirs} & \emph{emself} \\
Nounself & \emph{vam} & \emph{vamp} & \emph{vamps} & \emph{vamps} & \emph{vampself}\\
Emojiself & \scalerel*{\includegraphics{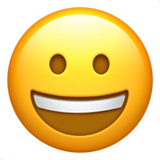}}{\strut} & \scalerel*{\includegraphics{img/grinning-face_1f600.png}}{\strut} & \scalerel*{\includegraphics{img/grinning-face_1f600.png}}{\strut}\emph{s}& \scalerel*{\includegraphics{img/grinning-face_1f600.png}}{\strut}\emph{s} & \scalerel*{\includegraphics{img/grinning-face_1f600.png}}{\strut}\emph{self} \\
Numberself & \emph{$0$} & \emph{$0$} & \emph{$0$s} &\emph{$0$s} &\emph{$0$self} \\
\bottomrule
\end{tabularx}
\caption{Phenomena and 3rd person pronoun sets by which they are represented in our analysis when translating from English (\textsc{en} $\rightarrow$ \textsc{da}, \textsc{de}, \textsc{fa}, \textsc{fr}, \textsc{it}). We list the pronouns for each grammatical case: nominative (N), accusative (A), possesive dependent (PD),  possessive independent (PI), and reflexive (R).}\label{tab:pronouns}
\end{table}

We fill the placeholders with pronouns of the correct grammatical case taken from 8 sets of pronouns that reflect diverse pronoun-related phenomena as described by \citet{lauscher2022welcome}. For example, we use \emph{she/ her /her/ hers/ herself} as an instance of gendered pronouns, and \emph{vam/ vamp / vamps/ vamps/ vampself} as an instance of nounself pronouns~\citep{miltersen2016nounself}. The latter are prototypically derived from a noun, and possibly match distinct aspects of an individual's identity. %
We list our test pronouns in Table~\ref{tab:pronouns}. 
Our setup allows us to test the translation of sentences containing different types of pronouns, in all of their grammatical forms, in more and less complex sentences and in contexts that are prone to different stereotypical associations.
Our procedure results in $164$ \textsc{en} sentences (4 sentences per 5 cases for each of the 8 pronoun sets plus 4 additional sentences for the variant \emph{themself} instead of \emph{themselves}).

\paragraph{Automatic Translation.} Next, we automatically translate the \textsc{en} source sentences to five languages: Danish (\textsc{da}), Farsi (\textsc{fa}), French (\textsc{fr}), German (\textsc{de}), and Italian (\textsc{it}). We choose these languages based on (a)~typological diversity, %
(b)~our access to native speakers, and (c)~their coverage by commercial MT. We ensure diversity with respect to family branches, scripts, and the handling of gender and pronouns in the languages: \textsc{de} and \textsc{da} represent the Germanic branch, \textsc{fr} and \textsc{it} the Romanic branch, and \textsc{fa} the Iranian branch of Indo-European languages. \textsc{da}, \textsc{de}, \textsc{fr}, and \textsc{it} employ the Latin script, and \textsc{fa} the Arabic one. Most importantly, the handling of grammatical gender and pronouns differs among languages. Concretely, \textsc{da}, \textsc{de}, \textsc{fr}, and \textsc{it} are gendered languages but differ in their number of genders (e.g., \textsc{de} has three grammatical genders while \textsc{fr} has two). While for  \textsc{de} and \textsc{it}, there is currently no gender-neutral pronoun recognized by an institutional body, for \textsc{fr}, the dictionary \emph{Le Robert} recently included the gender-neutral pronoun \emph{``iel''}. %
In contrast, \textsc{fa} is a gender-neutral language. Thus, there should also be no potential for misgendering in the resulting translations. Another interesting aspect is that \al{two} of the languages fall under the class of \emph{pro-drop} languages (\textsc{it}, \textsc{fa})\footnote{
\final{Pro-drop refers to a linguistic phenomenon where the subject pronoun can be omitted from a sentence without affecting its grammaticality or clarity. It is often clear from the verb inflection, as in Italian \emph{``Vado'': ``(I) go.''}}}, while the others do not allow for dropping the pronoun.

We focus on assessing the state of commercial MT, and accordingly rely on 3 established MT engines: Google Translate,\footnote{\url{https://translate.google.com}; we accessed Google Translate through the interface provided in Google Sheets. Note that we observed differences in translation when using the graphical user interface.} Microsoft Bing,\footnote{\url{https://www.bing.com/translator}}  and DeepL Translator.\footnote{\url{https://www.deepl.com/translator}} Currently, DeepL does not cover Farsi (all other languages are covered by all three commercial MT engines).

\paragraph{Annotation Criteria.} While initially, we wanted to focus solely on identity aspects conveyed by the pronouns, we noticed in an early pre-study that some of the translations exhibited more fundamental issues. This is why we resort to the following three categories, which allow us to answer research questions RQ1--RQ3, to guide our analysis of a translation B based on an \textsc{en} sentence A: \emph{grammatical correctness}, \emph{semantic consistency}, and \emph{pronoun translation behavior}.

\vspace{0.3em}
\noindent\emph{(1) Grammatical Correctness.} We ask our annotators to assess whether translation B is grammatically correct. Annotators are instructed to not let their judgment be affected by the occurrence of neopronouns that are potentially uncommon in the target language, e.g., emojiself pronouns. 

\vspace{0.3em}
\noindent\emph{(2) Semantic Consistency.} We let our annotators judge whether B conveys the same message as A in two variants: First, we seek to understand whether \emph{independent} of how the pronoun was translated the semantics of A are preserved. Second, we ask whether when also considering the pronoun translation, semantics are preserved.

\vspace{0.3em}
\noindent\emph{(3) Pronoun Translation Behavior.} The third category specifically focuses on assessing the translation of the pronoun. We investigate whether the pronoun was \emph{omitted} (i.e., it is not present in B), \emph{copied} (pronoun in B is exactly the same as in A), or \emph{translated} (the system output some other string in B as correspondence to the pronoun in A). %
\final{Note that none of these cases necessarily corresponds to a translation error (or translation success) -- for instance, it might be a valid option to directly copy the pronoun from the input in the source language to fully preserve its individual semantics. }
If the pronoun was ``translated'', we ask annotators to highlight its translation, and to further indicate if the translation corresponds to a common pronoun in the target language (and also, whether it still functions as a pronoun). If a common pronoun is chosen, we also collect its number and its commonly associated gender.

\paragraph{Annotation Process.} As the evaluation task requires annotators to be familiar with the target language, the concept of neopronouns, and linguistic properties such as part-of-speech tags, we hired five native speakers of target languages who all hold a university degree, are proficient speakers of English, and have diverse gender identities (man, woman, non-binary). \final{We payed our annotators 15€ per hour, which is substantially above the minimum wage in Italy and in line with the main authors' university recommendations for academic assistants.}

All annotators demonstrated great interest in helping to make MT more inclusive and were familiar with the overall topic. We took a descriptive annotation approach \cite{roettger-etal-2022-two}. Each annotator then underwent specific training in 1:1 sessions in which we showed them examples and offered room for discussions and questions. 
To facilitate the task and guide our annotators through the annotation criteria, we developed a specific annotation interface (see \al{Appendix}). To assess the reliability of our evaluation, we hired a second annotator for \textsc{de} and \textsc{it} to compute inter-annotator agreement and let the same native speaker of \textsc{fa} re-annotate a portion of the data to compute intra-annotator agreement (50 instances each). We measured an inter-annotator-agreement (Krippendorff's $\alpha$) of 0.73 for \textsc{de} and 0.69 for \textsc{it}, and an \textit{intra}-annotator agreement \cite{abercrombie2023consistency} of 0.86 for \textsc{fa} across all upper-level categories. We thus assume our conclusions to be valid. After completing the assessment, we gave every worker access to their annotations with the option to change and clean their results. %

\subsection{Results}
\begin{figure*}[t]
     \centering
    \begin{subfigure}[b]{0.329\textwidth}
         \centering
         \includegraphics[width=\linewidth, trim=2em 0em 2.5em 2em, clip]{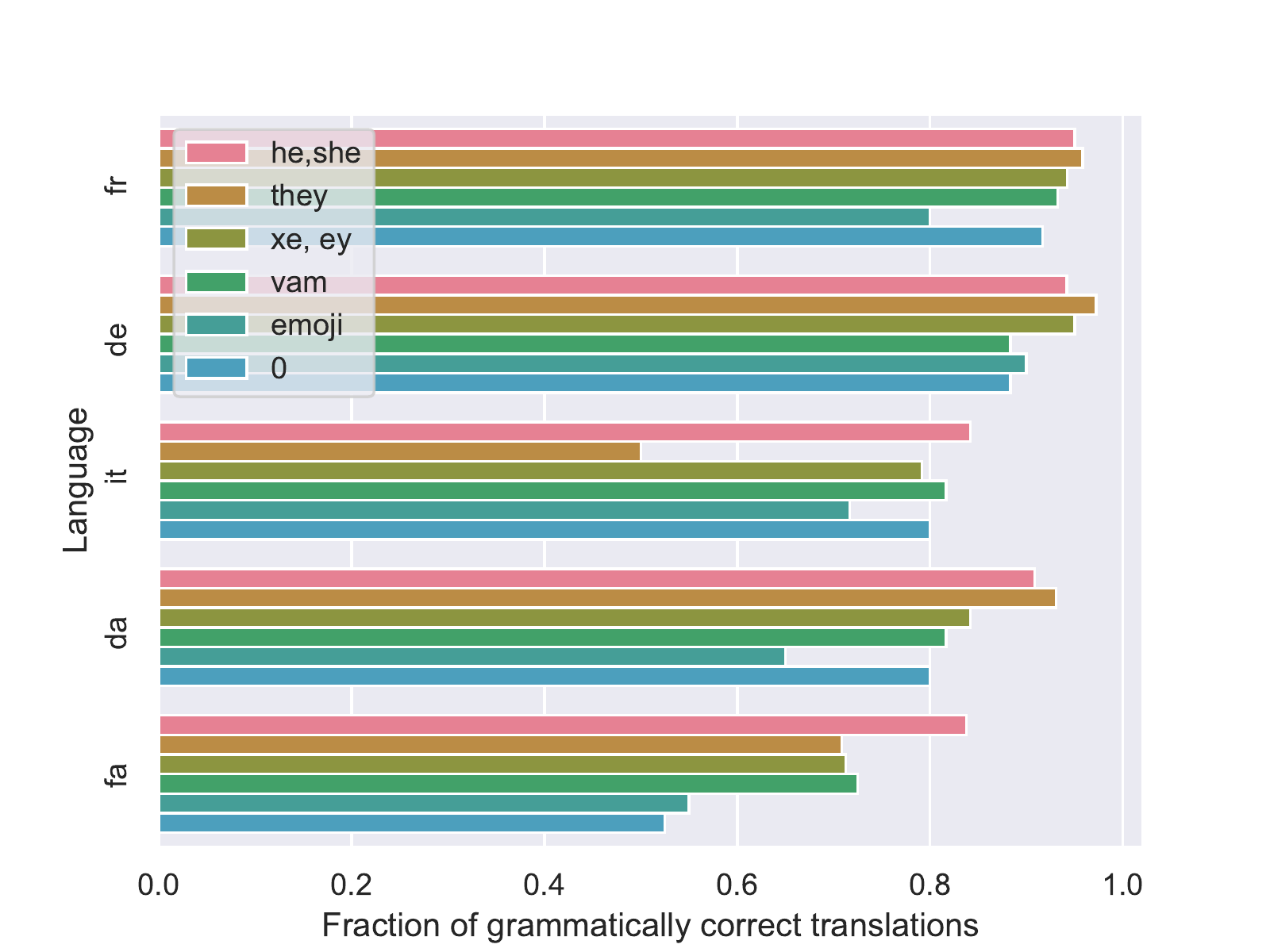}
         \caption{Grammaticality}
         \label{fig:overall_grammaticality}
     \end{subfigure}
     \hfill
     \begin{subfigure}[b]{0.329\textwidth}
         \centering
         \includegraphics[width=\linewidth, trim=2em 0em 2.5em 2em, clip]{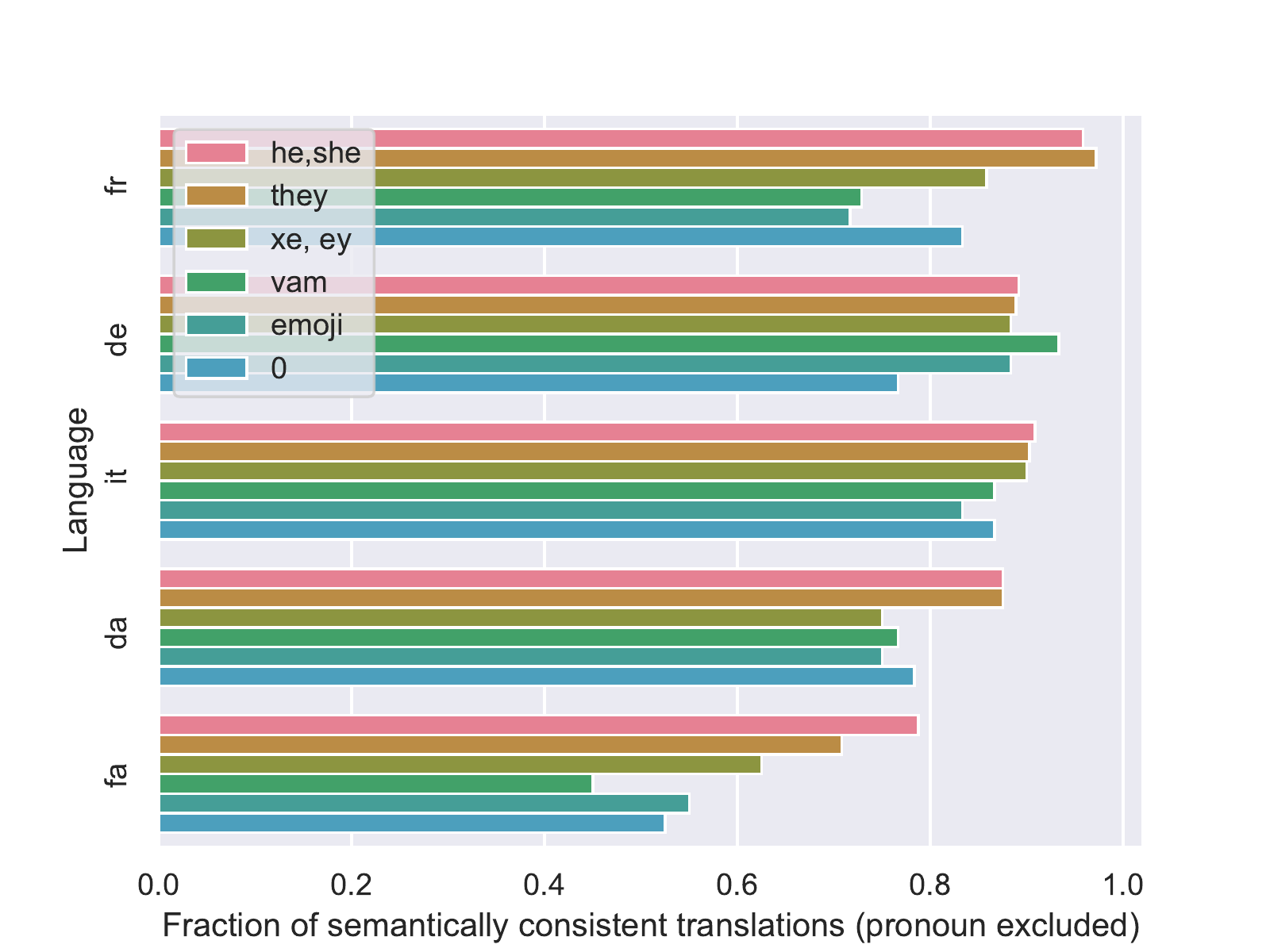}
         \caption{Semantics: pronoun excluded}
         \label{fig:overall_semantics_excluded}
     \end{subfigure}
     \hfill
    \begin{subfigure}[b]{0.329\textwidth}
         \centering
         \includegraphics[width=\linewidth, trim=2em 0em 2.5em 2em, clip]{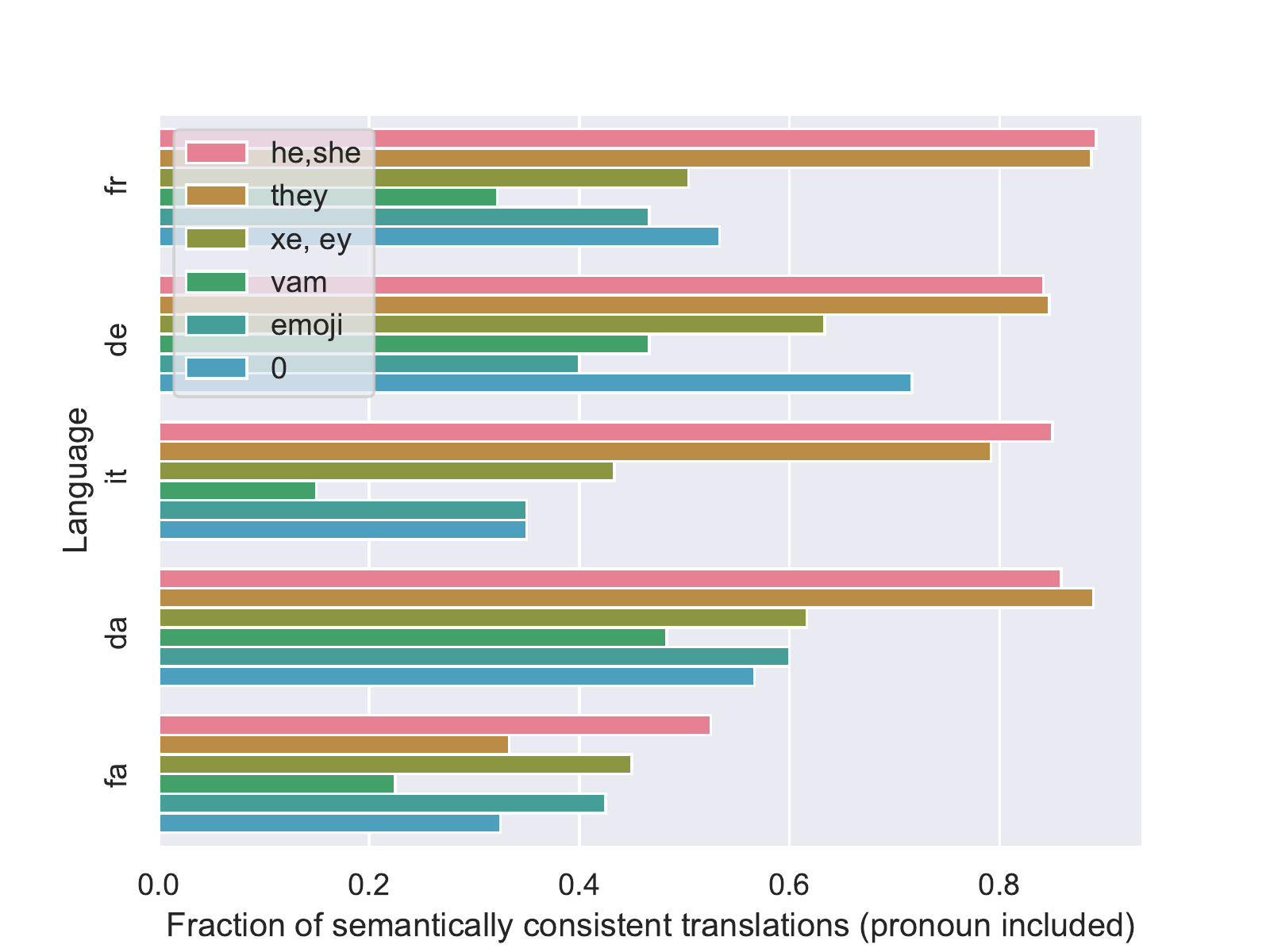}
         \caption{Semantics: pronoun included}
         \label{fig:overall_semantics_included}
     \end{subfigure}
    \caption{Overall translation quality. We show the fraction (\%) of grammatically correct (a) and  semantically correct (pronoun excluded (b) or included (c)) translations aggregated across all three engines for five target languages (Danish (\textsc{da}), Farsi (\textsc{fa}), French (\textsc{fr}), German (\textsc{de}), Italian (\textsc{it})) given English input sentences containing the following pronoun groups: he, she (gendered); they (gender-neutral); xe, ey (``traditional'' neopronouns); vam (nounself pronoun); \scalerel*{\includegraphics{img/grinning-face_1f600.png}}{\strut} (emojiself pronoun); and 0 (numberself pronoun).}
\end{figure*}
\renewcommand{\arraystretch}{1.3}
\setlength{\tabcolsep}{4pt}
\begin{table*}[!t]\centering
\small
\begin{tabularx}{\linewidth}{r X X l  l}\toprule
& \textbf{Source} & \textbf{MT Output} & \textbf{Case} &  \textbf{Issue Type} \\\midrule

1 & \emph{The attendant did not want to fight with the guard and gave \textbf{them} flowers.} & \emph{Die Wärterin wollte sich nicht mit dem Wachmann streiten und schenkte \textbf{ihm} Blumen.} & Accusative  & Potential Misgendering \\

2 & \emph{Liam received a good grade, so \textbf{vam} was happy.} & \emph{Liam erhielt eine gute Note, und \textbf{Vam} war zufrieden.} & Nominative  & Semantic Inconsistency \\

3 & \emph{Olivia found \textbf{eir} ring.} & \emph{Olivia fand \textbf{einen Eir-Ring}.} & Poss. Depen.  & Semantic Inconsistency \\

4 & \noindent\emph{During lunch, the janitor looked for the attendant to steal \textbf{eirs}.} & \emph{Während des Mittagessens suchte der Hausmeister nach dem Besucher, um \textbf{Eurren} zu stehlen.} &  Poss. Indep.  &  Pronoun Mistranslation \\

5 & \emph{Liam wanted a new haircut, so \scalerel*{\includegraphics{img/grinning-face_1f600.png}}{\strut} cut the hair \scalerel*{\includegraphics{img/grinning-face_1f600.png}}{\strut}self.} & \emph{Liam wollte einen neuen Haarschnitt, also schneiden \textbf{Sie} das Haar \scalerel*{\includegraphics{img/grinning-face_1f600.png}}{\strut} \textbf{selbst}.} &Reflexive  & Semantic Inconsistency \\

\bottomrule
\end{tabularx}
\caption{\final{Problems in the  MT output. We show examples for \textsc{en} to \textsc{de} for different grammatical cases, pronouns, and issue types, and highlight the pronouns in the source sentence and the corresponding parts in the translation in bold.}}\label{tab:mistranslations}
\end{table*}
\paragraph{Overall translation quality.} We show the results on grammaticality and semantic consistency in Figures~\ref{fig:overall_grammaticality}--\ref{fig:overall_semantics_included}. Depending on the target language as well as the pronoun category, the performance varies greatly; for instance, while for gendered pronouns in \textsc{fr} 95\% of the translations are grammatically correct, we observe a drop of 15 percentage points for emoji-self pronouns. Even more severely, only half (!) of the translations to \textsc{it} are grammatically correct when starting with the gender-neutral pronoun \emph{``they''} (Figure~\ref{fig:overall_grammaticality}). We make similar observations when asking annotators whether the meaning is preserved during the translation process (semantic consistency): Even when not considering the translation of the pronoun, in most cases, the performance drops when moving from a gendered to a gender-neutral pronoun set. We note the biggest drop, 34 percentage points, for \textsc{fa} and the category of noun-self pronouns (45\% ) compared to gendered pronouns with 79\% (Figure~\ref{fig:overall_semantics_excluded}). Compared to the results for gendered pronouns, we note the following maximum drops when aggregating over all languages we test: 16 percentage points for grammaticality, 13 percentage points for semantic consistency (pronoun excluded), both towards emoji-self pronouns, and a huge drop of 47 percentage points for semantic consistency when the pronoun is included in the assessment. We provide the aggregated plots in the \al{Appendix}. 
\begin{figure*}[t]
     \centering
    \begin{subfigure}[b]{0.329\textwidth}
         \centering
         \includegraphics[width=\linewidth, trim=0.5em 0em 1.5em 0.5em, clip]{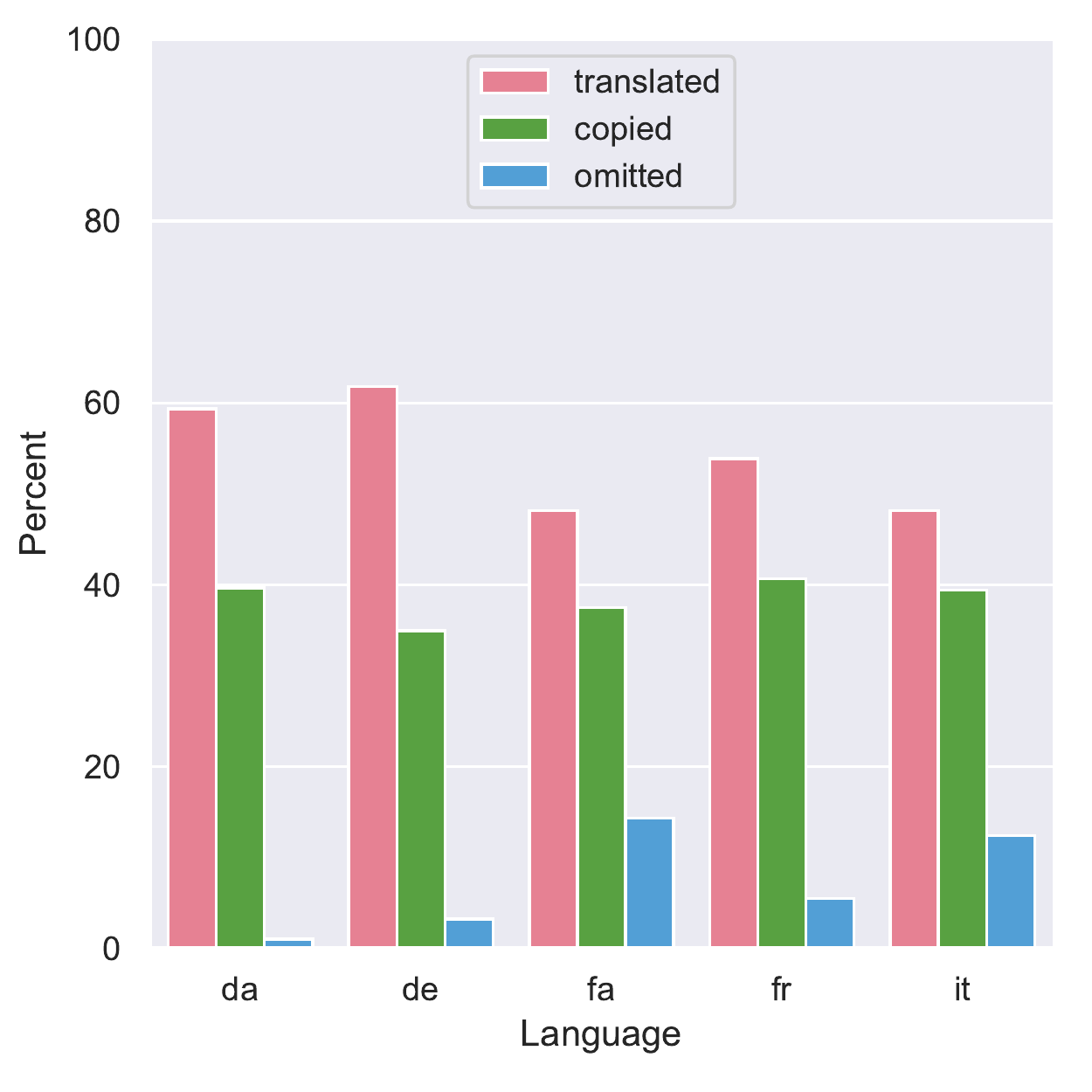}
         \caption{Language}
         \label{fig:treatment-langs}
     \end{subfigure}
     \hfill
     \begin{subfigure}[b]{0.329\textwidth}
         \centering
         \includegraphics[width=\linewidth, trim=0.5em 0em 1.5em 0.5em, clip]{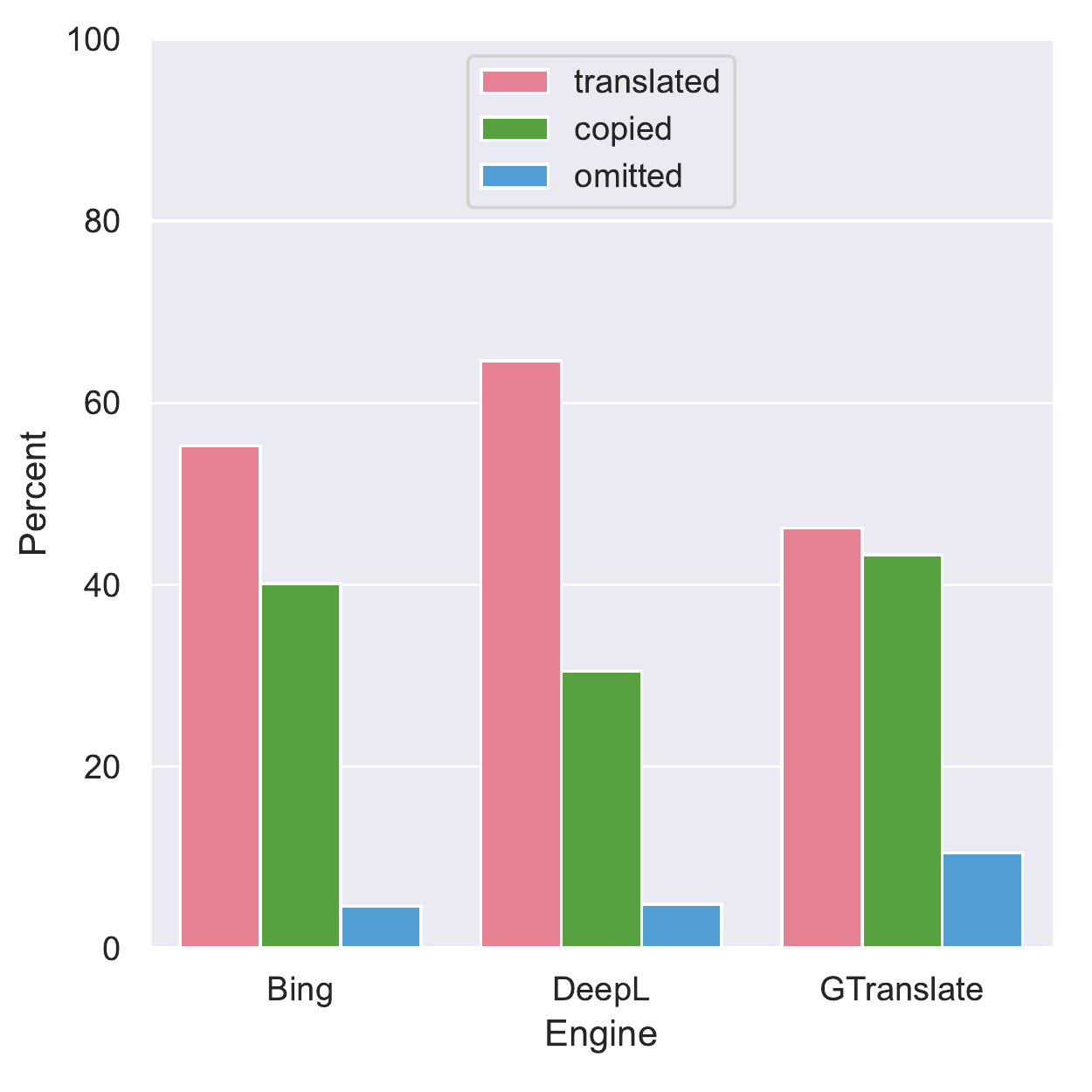}
         \caption{Translation engine}
         \label{fig:treatment-engines}
     \end{subfigure}
     \hfill
    \begin{subfigure}[b]{0.329\textwidth}
         \centering
         \includegraphics[width=\linewidth, trim=0.5em 0em 1.5em 0.5em, clip]{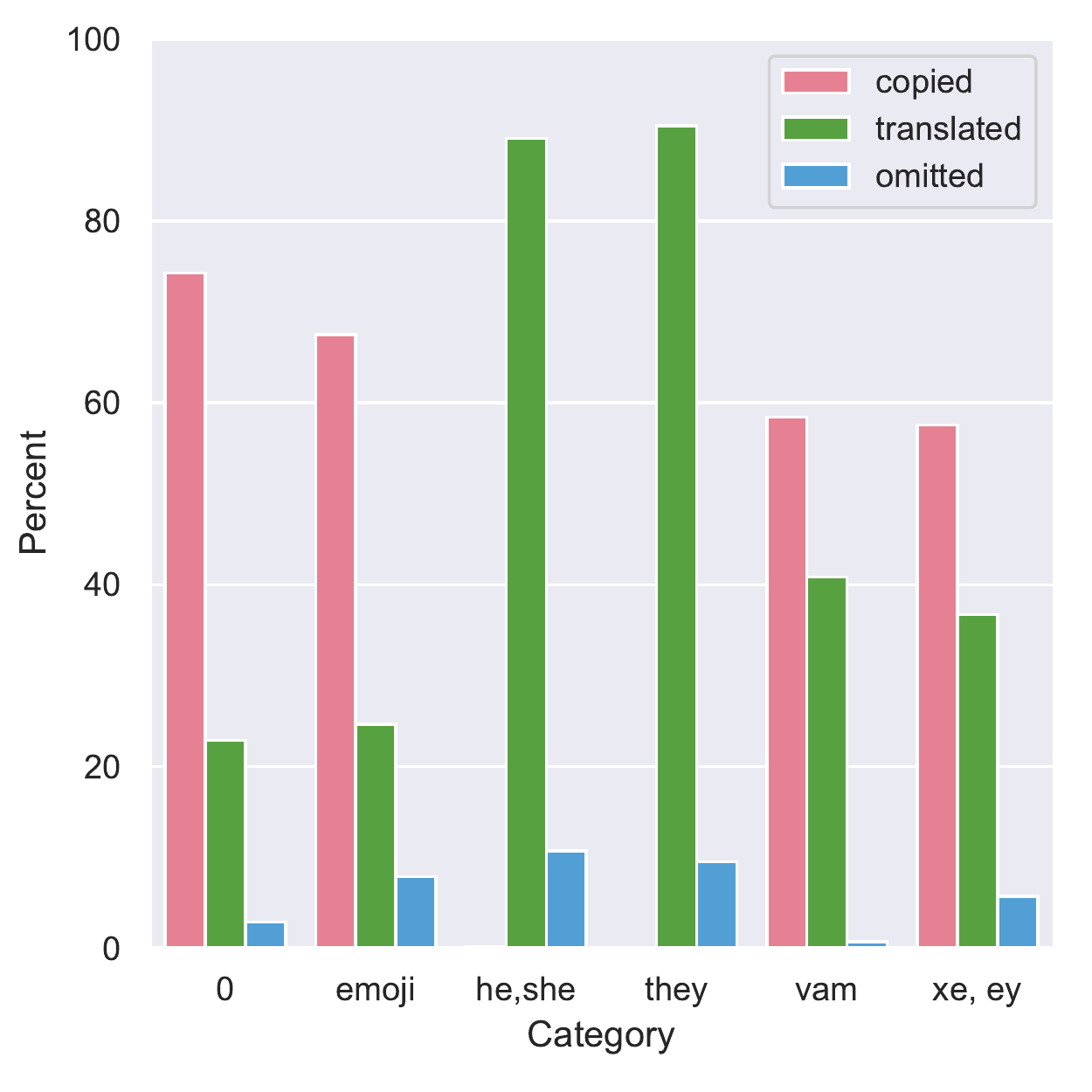}
         \caption{Pronoun group}
         \label{fig:treatment-group}
     \end{subfigure}
    \caption{Pronoun treatment strategies. We show the fraction (\%) of translated, copied, and omitted pronouns (a) per language (French (\textsc{fr}), German (\textsc{de}), Italian (\textsc{it}), Danish (\textsc{da}), Farsi (\textsc{fa})), (b) per translation engine, and (c) per pronoun group in the English input sentence (\emph{he, she} (gendered); \emph{they} (gender-neutral); \emph{xe, ey} (``traditional'' neopronouns); \emph{vam} (nounself pronoun); \scalerel*{\includegraphics{img/grinning-face_1f600.png}}{\strut} (emojiself pronoun); and \emph{0} (numberself pronoun)).}
\end{figure*}

\paragraph{Pronoun treatment strategies.} We depict the different strategies of how pronouns are treated in the translation in Figures~\ref{fig:treatment-langs}--\ref{fig:treatment-group}. Across all languages, the engines most often ``translate'' the pronouns (up to $\sim$62\% for \textsc{de}), i.e., some non-identical string corresponding to the \textsc{en} input pronoun is present in the output. The most unpopular strategy is to omit the pronoun. Unsurprisingly, the highest fraction of translations where this strategy is applied is present among the pro-drop languages, \textsc{fa} (14\%) and \textsc{it} (12\%). Among the three translation engines, DeepL exhibits the highest fraction of pronoun translations (65\%).\footnote{Note, however, that \textsc{fa} is not included due to coverage.} In contrast, GTranslate is the engine with the largest pronoun copies (43\%). Interestingly, we again observe a huge variation among the different pronoun groups: while the gendered pronouns (\emph{he}, \emph{she}) and the gender-neutral pronoun (\emph{they}) are most often translated (89\% and 90\%, respectively) and are almost never copied to the output, our representatives of the number-self and emoji-self pronouns most often are (74\% and 68\%, respectively). This is also the case for the noun-self pronoun (\emph{vam}) and the more traditional neopronouns %
(\emph{xe}, \emph{ey}), with roughly 58\% of copies each. However, for these, the fraction of translations in turn greatly surpasses those of numberself and emojiself pronouns, with 41\% and 37\%. %

\begin{figure}[t]
     \centering
     \includegraphics[width=\linewidth, trim=0em 2.2em 0.0em 0em, clip]{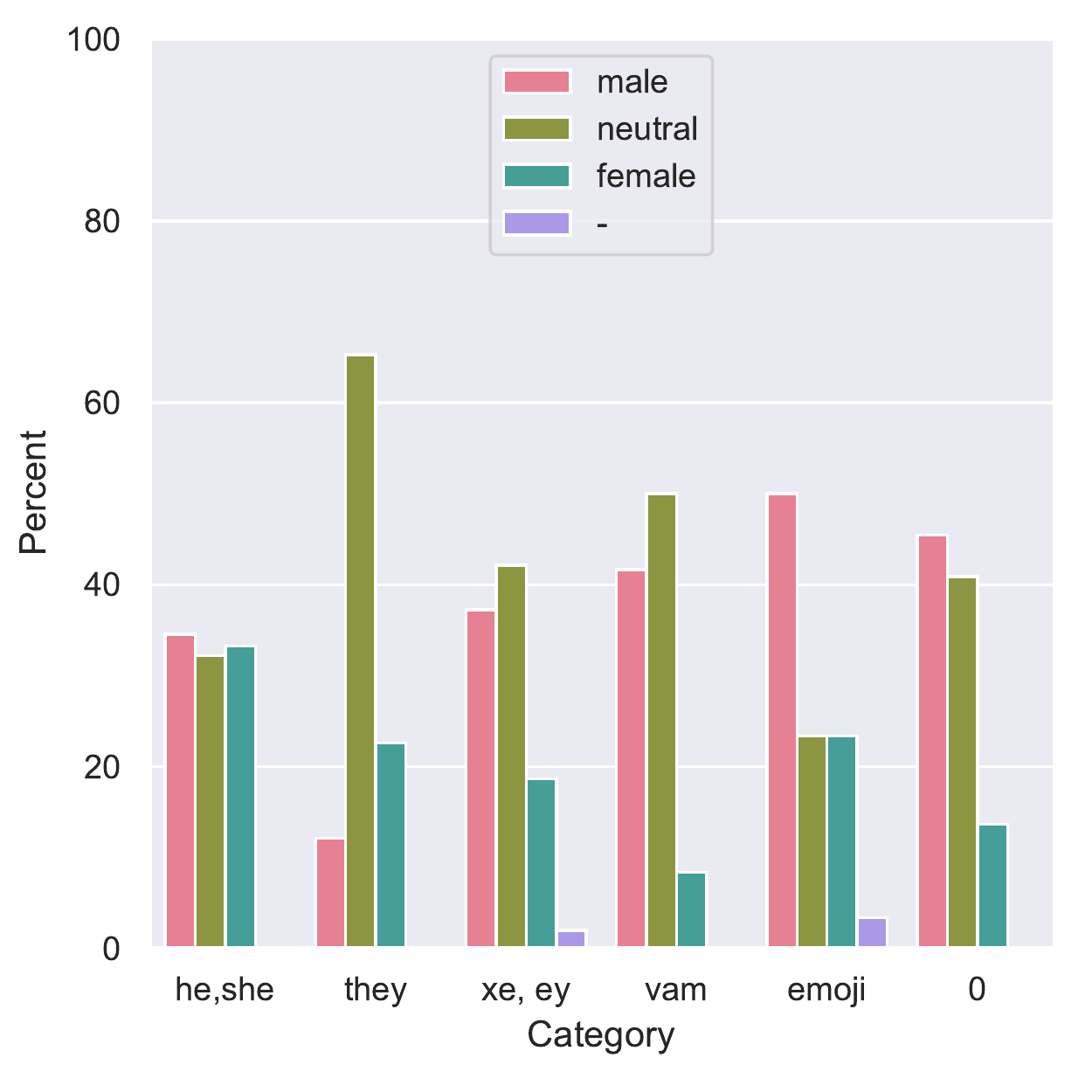}
    \caption{Gender conveyed by the target language pronoun (\emph{male}, \emph{neutral}, \emph{female}, \emph{unknown (--)}) for translations that contain an existing third-person singular pronoun. We aggregate across languages. For Italian and French, we focus on the gender of the subject. \final{We exclude Farsi due to its gender neutrality.}
    }
    \label{fig:translated_pronoun_gender}
\end{figure}
\paragraph{Translation and Gender.} We analyze pronouns that are translated to an existing singular pronoun in the target language in Figure~\ref{fig:translated_pronoun_gender}. For the gendered source pronouns (\emph{he, she}), the result is roughly balanced across commonly associated genders. For \emph{they}, we observe a high proportion of gender-neutral output pronouns (65\%)---most often, gender neutrality is preserved. In contrast, for different types of neopronouns, the engines are likely to output a gendered pronoun. This finding is most pronounced for emojiself pronouns, with 50\% and 23\% of output pronouns commonly associated with male and female individuals, respectively. This amount of translations (73\%) is likely to correspond to cases of misgendering.

\final{\paragraph{Qualitative Analysis.} For further illustration, we show examples of some problems we observe when translating to \textsc{de} in Table~\ref{tab:mistranslations}. The output in Example 1 is generally correct. However, the gender-neutral pronoun \emph{they} is translated to the gendered pronoun \emph{er}. Examples 2 and 3 show translations in which the pronoun correspondence is copied from the input but starts with a capital letter (or is even prepended to the succeeding word, e.g., \emph{Eir-Ring}), as done for nouns or names. We note a similar problem in example 4. Additionally, the output string corresponding to the pronoun is neither copied from the input nor corresponds to a valid word in the target language (\emph{Eurren}). Finally, in example 4, the emojiself pronoun appears in the output translation with the additional 2nd person pronoun variant \emph{Sie}.} 

\subsection{Translating to English}
\label{sec:to_en}
\paragraph{Experimental Setup.} So far, we have started from \textsc{en} source sentences. Here, we expand our perspective and conduct the inverse experiment: We translate \emph{to} \textsc{en} starting from \textsc{da} sentences (as an example of a language with a recently emerging gender-neutral pronoun). To this end, we start from our \textsc{en} templates and manually translate these to \textsc{da}. We then fill the templates with the pronouns \emph{han} (=\emph{he}), \emph{hun} (=\emph{she}), \emph{hen} (gender-neutral), resulting in 48 source sentences. We translate those automatically with the three commercial engines and let an English native speaker evaluate the output according to the same guidelines.  

\paragraph{Results.} The overall translation quality is relatively high; for instance, we find that 75\% of translations are grammatically correct when starting from the gendered pronouns (\emph{han, hun}), and only see a small drop for the gender-neutral pronoun (\emph{hen} with 71\%). However, surprisingly, the translation engines seem to never output the gender-neutral option \emph{``they''} when choosing an existing pronoun in the target language \textsc{en}, not even when starting from \emph{hen}. In contrast, in roughly 72\% of the cases, \emph{hen} is translated to \emph{he}.

%% file: 05-survey.tex
Our results show that %
commercial engines cannot deal with pronouns as an open word class. Often, the output is not grammatical, and the meaning is inconsistent. %
Beyond these general aspects %
we have shown that %
pronoun treatment strategies vary. Next, we seek to understand how %
individuals would want their pronouns to be handled (RQ4).
\setlength{\tabcolsep}{7pt}
\begin{table}[t]
    \centering
    \small{
    \begin{tabular}{lrl}
        \toprule
         \textbf{Lang.} & \textbf{\% Ment.}  & \textbf{Pronoun sets} \\
         \midrule
        \textsc{de} & 35.00 & \emph{er, sie, dey, ey, <none>}\\
         \textsc{en} & 30.00 & \emph{he, she, they, it, <no preference>}\\
         \textsc{da} & 20.00 & \emph{han, hun, den, de, she, they}\\
         \textsc{it} & 7.50 & \emph{lei, lui}\\
         \textsc{ru} & 5.00 &\emph{\foreignlanguage{russian}{он}, <none>} \\
         \textsc{fa} & 2.50 & \emph{\foreignlanguage{farsi}{او}}\\
\bottomrule
    \end{tabular}}
    \caption{Fraction of mentions (in \%) of native languages (Danish (\textsc{da}), English (\textsc{en}), German (\textsc{de}), Italian (\textsc{it}), Russian (\textsc{ru}), and Farsi (\textsc{fa})) with associated pronouns participants of our survey identify with.}
    \label{tab:langs}
\end{table}
\begin{figure*}[t!]
    \centering
    \includegraphics[width=\textwidth]{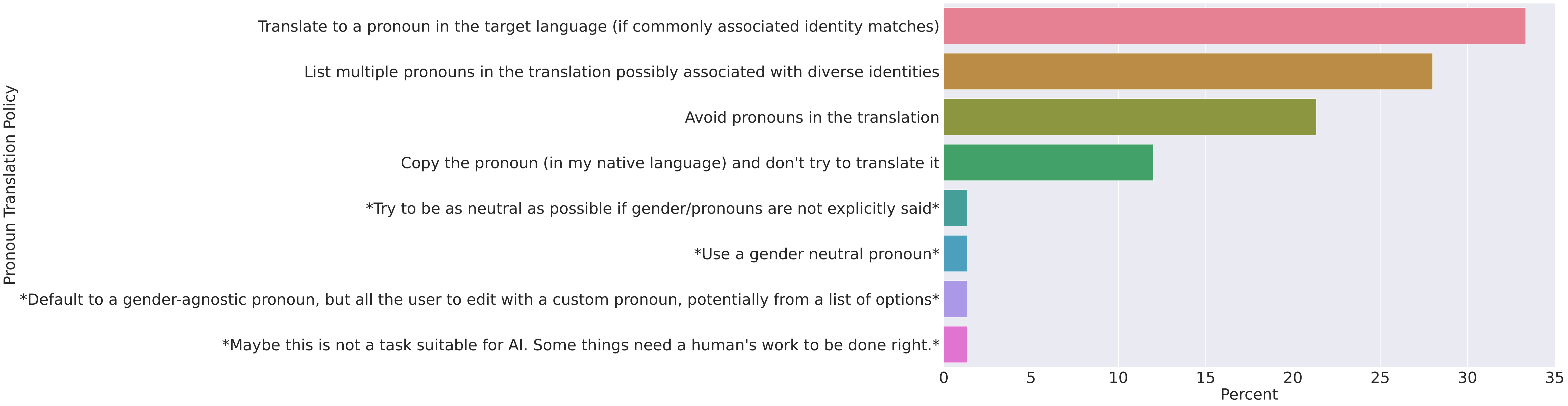}
    \caption{Results for our survey question relating to MT pronoun policies. Answers in asterisks (*) were additionally provided by our participants. We show them here for completeness.}
    \label{fig:policies}
\end{figure*}
\setlength{\tabcolsep}{16pt}
\begin{table*}[t!]
    \centering
    \small{
    \begin{tabular}{ll}
        \toprule
         \textbf{Translation policy} & \textbf{Translation} \\
         \midrule
         Referent's name & \emph{Liam hat eine gute Note bekommen, also war \textbf{Liam} glücklich.} \\
         Ellipsis through alternative construct & \emph{Liam erhielt eine gute Note und war deshalb froh.}\\
         General noun (person) & \emph{Liam hat eine gute Note bekommen, deshalb war \textbf{die Person} glücklich.} \\
         Neopronoun & \emph{Liam hat eine gute Note bekommen, deswegen freut \textbf{dey} sich.} \\
\bottomrule
    \end{tabular}}
    \caption{German example translations for the English source sentence \emph{``Liam received a good grade, so \textbf{they} were happy.''} provided by native speakers. Participants choose various policies for preserving gender neutrality.}
    \label{tab:translations}
\end{table*}
\subsection{A Survey on Pronouns and MT}
\paragraph{Survey Design and Distribution.} 
To answer this RQ, we design a survey consisting of three parts: 
(1) a general part asks for the participant's demographic information, e.g., age, (gender) identity, as well as their pronouns in English and their native languages. 
(2) The second part asks general opinions related to  pronouns in artificial intelligence. %
(3)~The last section deals specifically with MT: here, we ask how the individual would like their or their friends' pronouns to be treated when translating from their native language to another. 

Participants can choose from four treatment options we identified through informal discussions with affected individuals: 
\emph{(a) Avoid pronouns in the translation}; 
\emph{(b) Copy the pronoun (in my native language) and don't try to translate it}; 
\emph{(c) Translate to a pronoun in the target language (if commonly associated identity matches)}; 
\emph{(d) List multiple pronouns in the translation possibly associated with diverse identities}. Participants can also define additional options.
We provide examples with gender-neutral pronouns in English and encourage the participant to provide a translation in their native language. The institutional review board of the main authors' university approved our study design. We distributed the survey through channels that allow us to target individuals potentially affected by the issue and who represent a wide variety of (gender) identities. Examples include QueerInAI,\footnote{\url{https://sites.google.com/view/queer-in-ai}}  %
 and local LGBTQIA+ groups, e.g., Transgender Network Switzerland.\footnote{\url{https://www.tgns.ch}} For validation, we ran a pre-study between March 22 and May 4, 2022 (with n=149). The main phase %
 was open for participation between June 18 and August 1, 2022.

\paragraph{Results.} 
In the main phase of our survey, 44 individuals participated. Their ages ranged from 14 to 43, with the majority between 20 and 30. For the analysis, we removed responses from participants under 18. The remaining participants provided diverse and sometimes multiple gender identity terms (e.g., \emph{non-binary, transgender, questioning, genderfluid}) and speak diverse native languages (e.g., English, German,  Persian). The fraction of mentions of native languages and provided pronoun sets per language are given in Table~\ref{tab:langs}: participants identify with diverse and sometimes multiple pronoun sets (e.g., gendered pronouns, neopronouns) as well as no pronouns. Interestingly, some seem to use \textsc{en} pronouns in their non-\textsc{en} native language. This observation aligns with the finding that bilinguals tend to code-switch to their L2 if it provides better options to describe their gender identity~\cite{kaplan2022binary}. In a similar vein, some participants provided only a gendered option in their native language (e.g., \emph{er} in German) but indicated to identify with a gender-neutral option in \textsc{en} (e.g., \emph{they}).

Concerning the translation policies, participants chose between 1 and 3 pre-defined options, and four provided additional ideas. The result is depicted in Figure~\ref{fig:policies}. While the most popular option is \emph{(c) Translate to a pronoun in the target language (if commonly associated identity matches)}, there is no clear consensus and also strong tendencies towards gender-agnostic solutions. This finding is supported by the example-based analysis where we asked participants to translate from English to their native language. Table~\ref{tab:translations} illustrates this finding via participant answers for English to German translations (German native speakers). Participants used different options, like using the referent's name or a neopronoun, to deal with the issue that there is no established gender-neutral pronoun in German.

Additional participant comments point to the difficulty of the problem, e.g., \emph{``this one's tough because it feels like different people are potentially going to have different desires on this one [...]''}. 
Overall, we thus conclude that \textbf{users' preferences are as diverse as the community itself}. 

\input{06-recommendation.tex}

%% file: 06-recommendation.tex
\subsection{Recommendations}
Based on our observations in \S\ref{sec:experiment} and the survey results, we provide three recommendations for making future MT more inclusive.

\vspace{0.3em}
\noindent\emph{(1) Consider pronouns an open word class when developing and testing MT systems.} As we have demonstrated, popular commercial MT systems often fail when gender-neutral pronoun sets are part of the input, even when translating between resource-rich languages like \textsc{en} and \textsc{it}. Thus, NLP researchers and practitioners must make MT more robust even with regards to fundamental properties such as grammaticality. Extending existing data sets to reflect a larger variety of pronouns is crucial.

\vspace{0.3em}
\noindent\emph{(2) If possible, provide options for personalization.} Our survey demonstrated no clear consensus on how pronouns should be treated, and that users' preferences and pronouns vary. Thus, if possible, i.e., if the user is aware of the pronouns referents in their input text identify with, and if they directly interact with the translation engine, the decision should be left to that user. This finding aligns with desideratum D5 for more identity-inclusive AI identified by \citet{lauscher2022welcome}.

\vspace{0.3em}
\noindent\emph{(3) Avoid potential misgendering as much as possible.} If options for personalization are limited, no translation strategy will be ideal for all users. However, instead of ``blindly'' translating, which, as we have demonstrated, is likely to lead to misgendering,  there are several other options that translation engines could choose that exhibit less potential for harm, e.g.,  gender-agnostic translations. %

%% file: 07-conclusion.tex
In this work, we have investigated the sensitivity of automatically translating pronouns: small words that can convey important identity aspects. To understand where current commercial MT stands with regards to this issue, we started with a thorough error analysis covering six languages and three MT engines. We demonstrated that the engines tested are more likely to produce low-quality output when starting from gender-neutral pronouns, and we further observed a high potential for misgendering. Emphasizing marginalized voices, we complemented our study with a survey of affected individuals. The answers led us to three recommendations for more inclusive MT. We hope our study will inform %
and fuel more research on these issues.   %

%% file: 0x-limitations.tex
Naturally, our work comes with a number of limitations: for instance, we restrict ourselves to testing eight pronoun sets out of the rich plethora of existing options. To ensure diversity, we resort to one or two sets per pronoun group---we hope that individuals feel represented by our choices. Similarly, we only translate single sentences and don't investigate translations of larger and possibly more complex texts and we only translate to a number of languages none of which is resource-lean. Our study demonstrates that simpler and shorter texts already exhibit fundamental problems in their translations, even to resource-rich languages. 

%% file: 0xx-ethics.tex
In this work, we present a reality check in which we show that established commercial MT systems struggle with the linguistic variety that is tied to the large spectrum of identities. Consequently, this work has an inherently ethical dimension: our intent is to point to the issue of subcultural exclusion in language technology. We acknowledge, however, that this issue is much bigger than the problems relating to the use of neopronouns and we hope to investigate the topic more globally in the future.

%% file: 0xxx-appendix.tex
\clearpage
\label{sec:appendix}
\section{Data Sets and Licenses}
In this work, we only made use of a single existing dataset, WinoMT\footnote{\url{https://github.com/gabrielStanovsky/mt_gender}}~\citep{stanovsky-etal-2019-evaluating}. We used the dataset to obtain \textsc{en} templates in different grammatical cases, which we filled with the pronouns we test. The data set is licensed under MIT License. We will publish our selection of sentences from WinoMT as well as the additional sentences we added under the same license.

\section{Additional Results}
\begin{figure*}[t]
     \centering
    \begin{subfigure}[b]{0.329\textwidth}
         \centering
         \includegraphics[width=\linewidth]{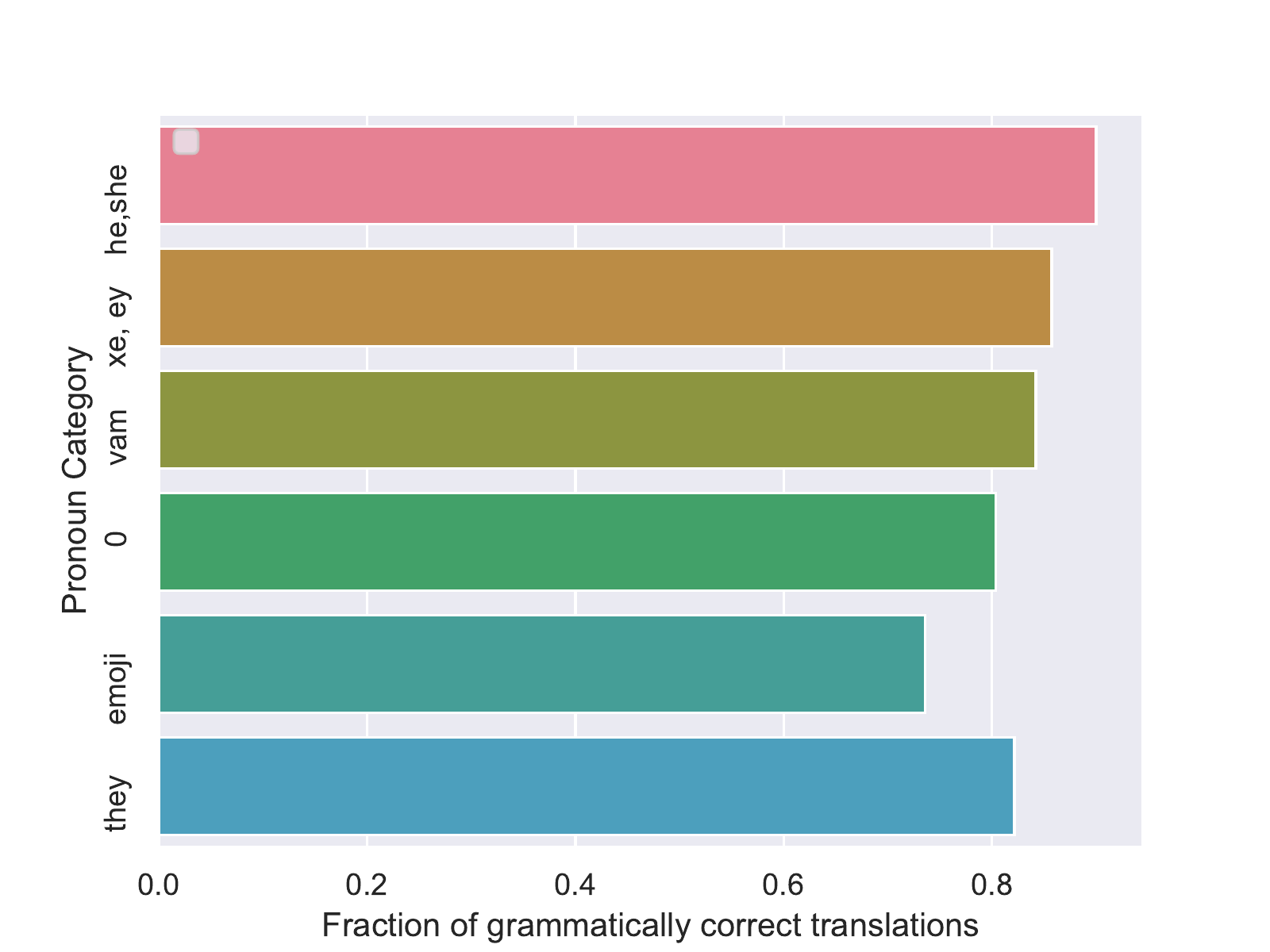}
         \caption{Grammaticality}
     \end{subfigure}
     \hfill
     \begin{subfigure}[b]{0.329\textwidth}
         \centering
         \includegraphics[width=\textwidth, clip]{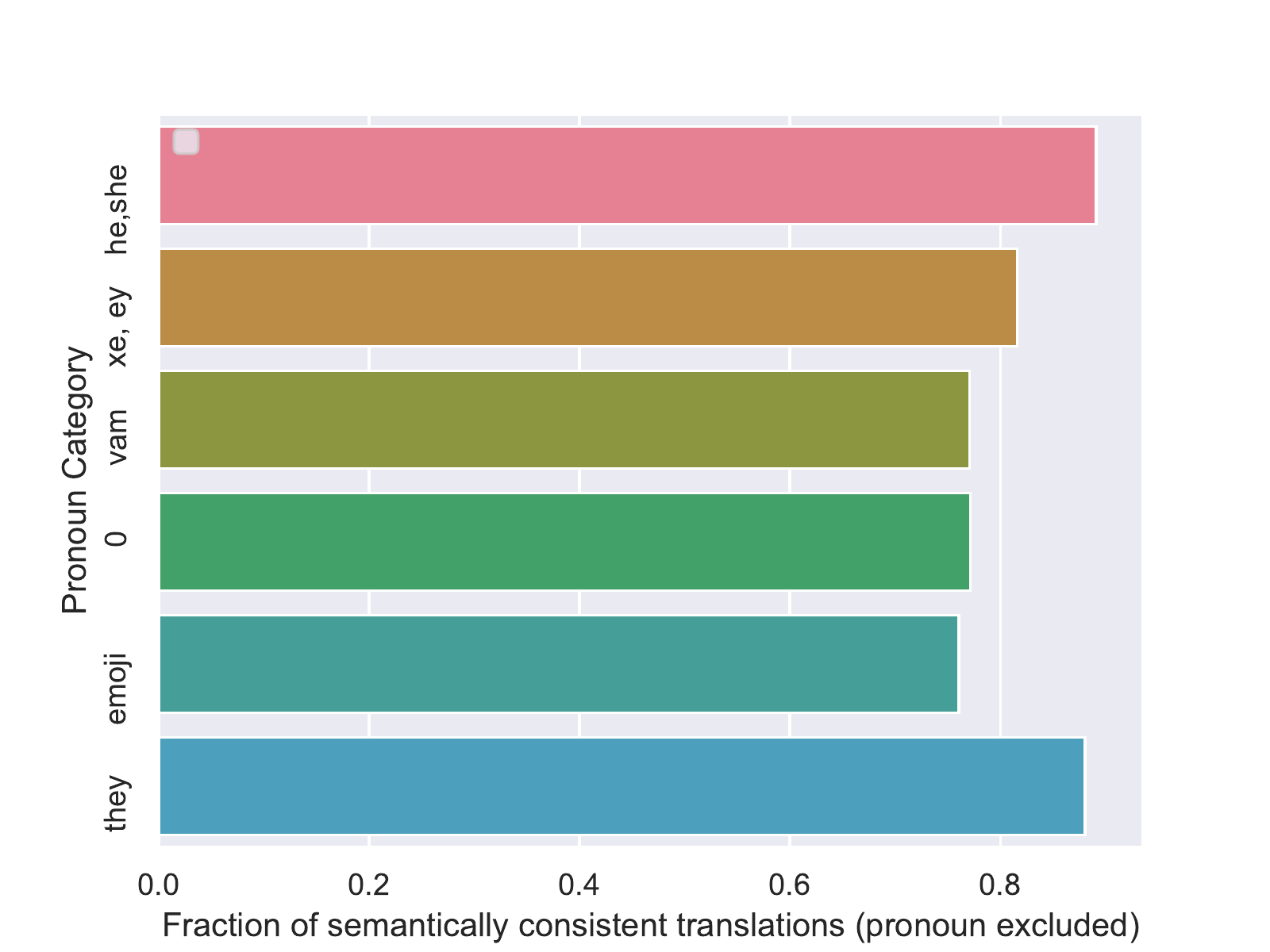}
         \caption{Semantics: pronoun excluded}
     \end{subfigure}
     \hfill
    \begin{subfigure}[b]{0.329\textwidth}
         \centering
         \includegraphics[width=\textwidth, clip]{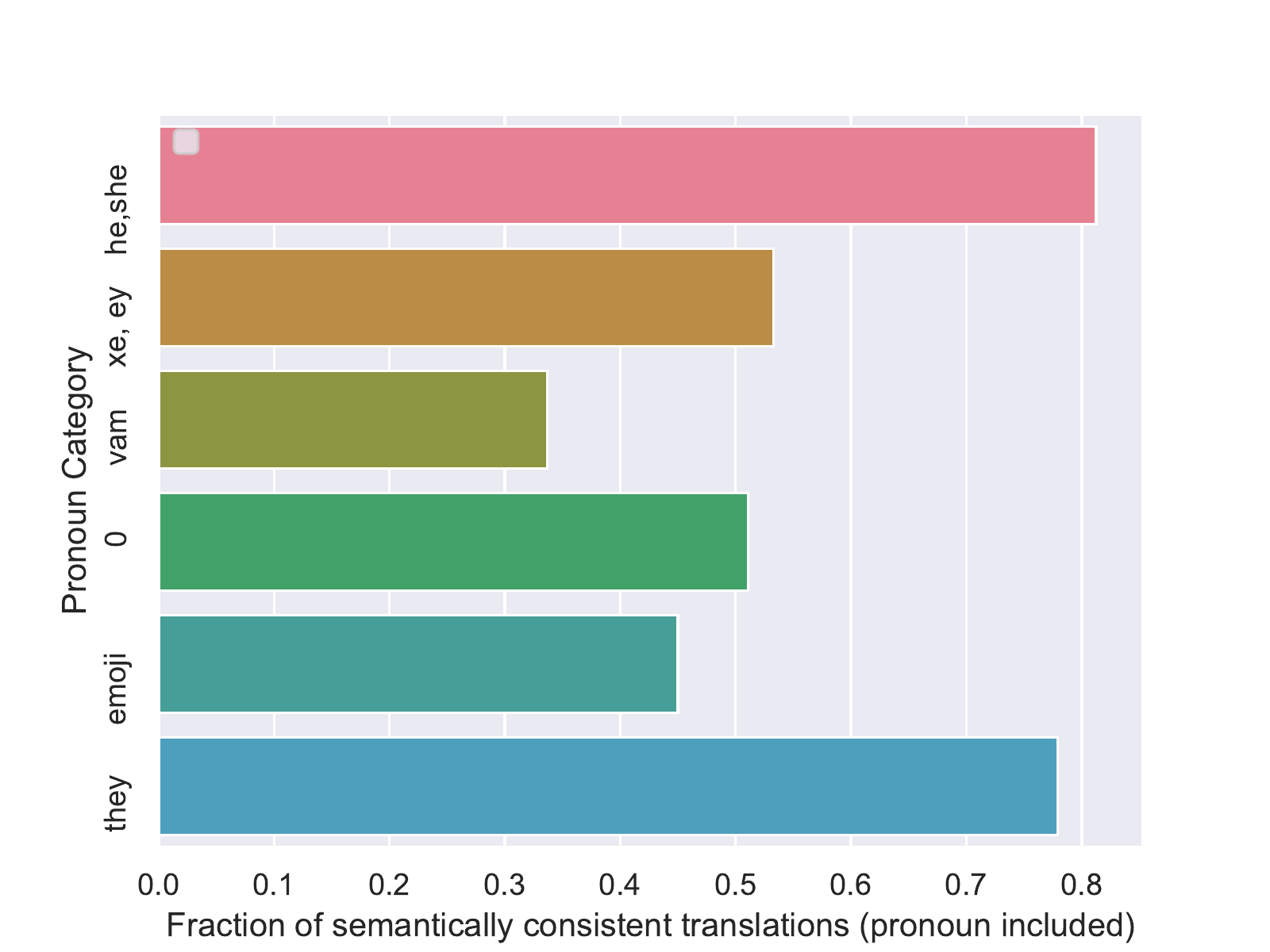}
         \caption{Semantics: pronoun included}
         
     \end{subfigure}
    \caption{Overall translation quality. We show the fraction (\%) of grammatically correct (a) and  semantically correct (pronoun excluded (b) or included (c)) translations aggregated across all three engines and five target languages given English input sentences containing the following pronoun groups: he, she (gendered); they (gender-neutral); xe, ey (``traditional'' neopronouns); vam (nounself pronoun); \scalerel*{\includegraphics{img/grinning-face_1f600.png}}{\strut} (emojiself pronoun); and 0 (numberself pronoun).}
    \label{fig:overall_across}
\end{figure*}

We provide additional results (aggregated across languages) in Figure~\ref{fig:overall_across}.

\section{Annotation Interface}
\begin{figure*}[t]
     \centering
    \begin{subfigure}[b]{1\textwidth}
         \centering
         \includegraphics[width=\linewidth]{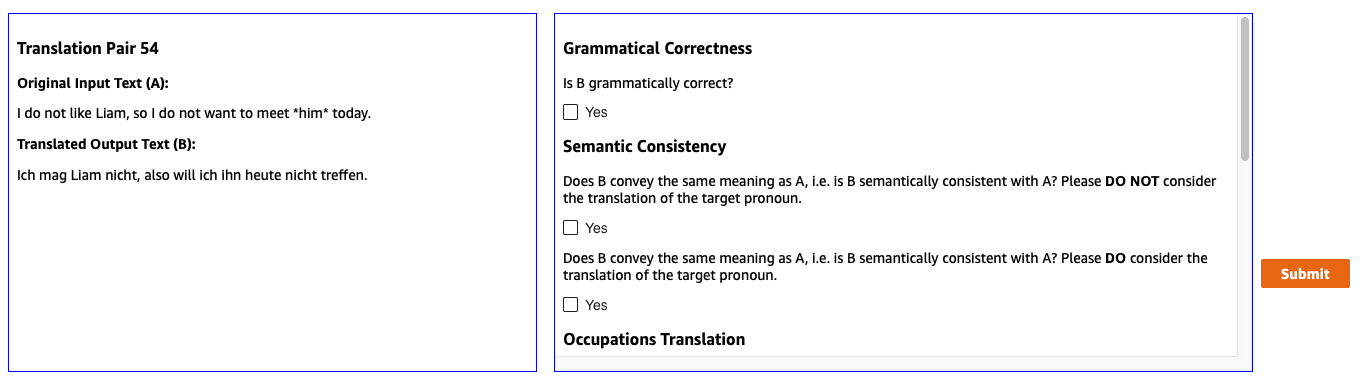}
         \caption{General Questions}

     \end{subfigure}
     \hfill
     \begin{subfigure}[b]{1\textwidth}
         \centering
         \includegraphics[width=\textwidth, clip]{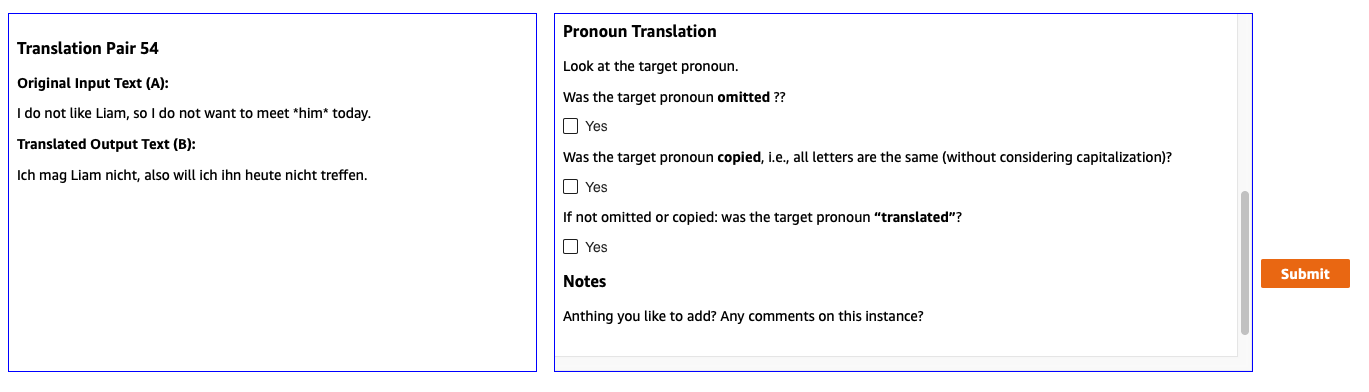}
         \caption{Pronoun Treatment}

     \end{subfigure}
     \hfill
    \begin{subfigure}[b]{1\textwidth}
         \centering
         \includegraphics[width=\textwidth, clip]{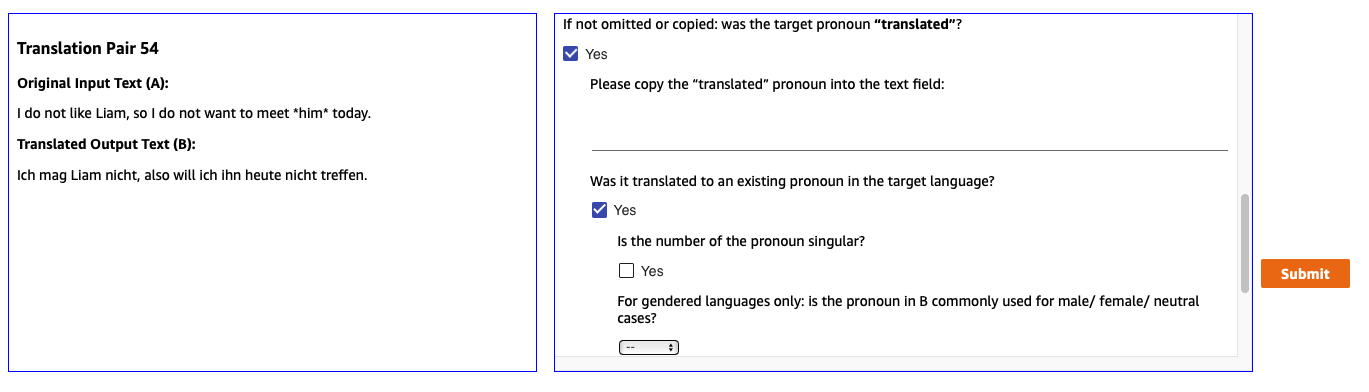}
         \caption{Translation Details}
     \end{subfigure}
    \caption{Screenshot of our annotation interface. The translation pair (here: \textsc{en} and \textsc{de}) is displayed on the left side of the screen. Annotators answer the questions shown on the right side: (a) general questions about the translation quality, (b) questions relating to whether the pronoun was ommitted, copied, or translated, and (c) details relating to the treatment strategy.}
    \label{fig:screenshot}
\end{figure*}

We show a screenshot of our annotation interface in Figure~\ref{fig:screenshot}. The interface was developed using HTML and JavaScript and hosted on the Amazon Mechanical Turk Sandbox.
